\documentclass[10pt,twocolumn,letterpaper]{article}

\usepackage{wacv}
\usepackage{times}
\usepackage{epsfig}
\usepackage{graphicx}
\usepackage{amsmath}
\usepackage{amssymb}
\usepackage{multirow}
\usepackage{float}
\newcommand{\thickhat}[1]{\mathbf{\hat{\text{$#1$}}}}

\renewcommand{\paragraph}[1]{\vspace{4pt}\noindent\textbf{#1}}

\def\wacvPaperID{283} % Enter the WACV Paper ID here

\wacvfinalcopy % *** Uncomment this line for the final submission

\ifwacvfinal
\def\assignedStartPage{1} % *** Enter the assigned starting page number (instead of 9876)
\fi

\ifwacvfinal
\usepackage[breaklinks=true,bookmarks=false]{hyperref}
\else
\usepackage[pagebackref=true,breaklinks=true,colorlinks,bookmarks=false]{hyperref}
\fi

\ifwacvfinal
\else
\fi

\begin{document}

%%%%%%%%% TITLE

\title{Rotate to Attend: Convolutional Triplet Attention Module}

\author{Diganta Misra \thanks{Equal Contribution}\\
Landskape\\
{\tt\small mishradiganta91@gmail.com}
\and
Trikay Nalamada \footnotemark[1]\\
Indian Institute of Technology, Guwahati\\
{\tt\small nalamada.trikay@gmail.com}
\and
Ajay Uppili Arasanipalai \footnotemark[1]\\
University of Illinois, Urbana Champaign\\
{\tt\small aua2@illinois.edu}
\and
Qibin Hou\\
National University of Singapore\\
{\tt\small andrewhoux@gmail.com}
}

\maketitle
%\thispagestyle{empty}

%%%%%%%%% ABSTRACT
\begin{abstract}
Benefiting from the capability of building inter-dependencies among channels or spatial locations, attention mechanisms have been extensively studied and broadly used in a variety of computer vision tasks recently. In this paper, we investigate light-weight but effective attention mechanisms and present triplet attention, a novel method for computing attention weights by capturing cross-dimension interaction using a three-branch structure. For an input tensor, triplet attention builds inter-dimensional dependencies by the rotation operation followed by residual transformations and encodes inter-channel and spatial information with negligible computational overhead. Our method is simple as well as efficient and can be easily plugged into classic backbone networks as an add-on module. We demonstrate the effectiveness of our method on various challenging tasks including image classification on ImageNet-1k and object detection on MSCOCO and PASCAL VOC datasets. Furthermore, we provide extensive in-sight into the performance of triplet attention by visually inspecting the GradCAM and GradCAM++ results. The empirical evaluation of our method supports our intuition on the importance of capturing dependencies across dimensions when computing attention weights. Code for this paper can be publicly accessed at \url{https://github.com/LandskapeAI/triplet-attention}

\end{abstract}

%%%%%%%%% BODY TEXT
\section{Introduction} \label{sec:introduction}

\begin{figure}
  \centering
%    \fbox{\rule{0pt}{2in} \rule{.9\linewidth}{0pt}}
  \includegraphics[width=0.46\textwidth]{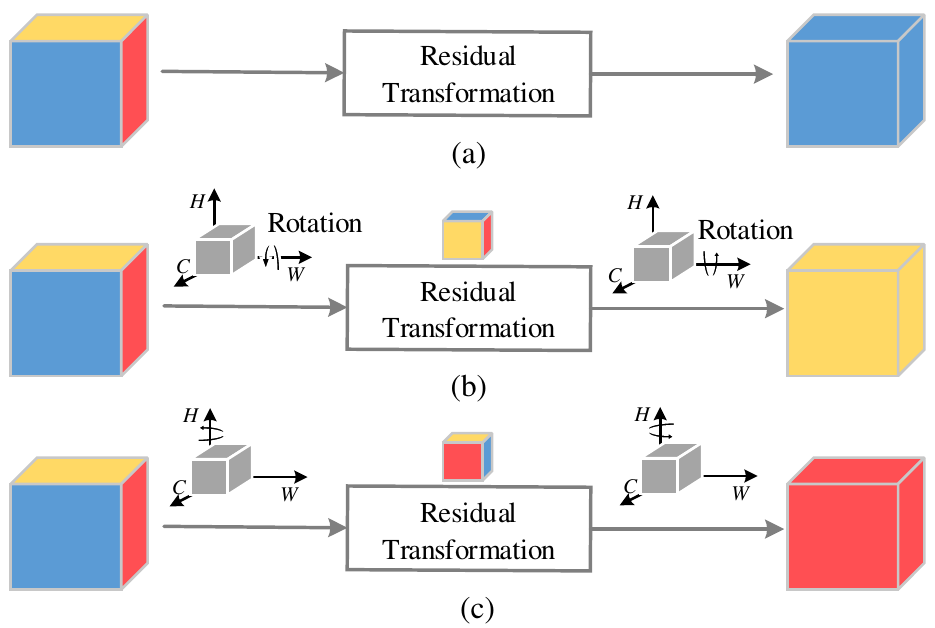} 
  \caption{Abstract representation of triplet attention with three branches capturing cross-dimension interaction. Given the input tensor, triplet attention captures inter-dimensional dependencies by rotating the input tensor followed by residual transformation.}
  \label{fig:teaser}
\end{figure}

Over the years of computer vision research,
convolutional neural network architectures of increasing depth
have demonstrated major 
success in many computer vision tasks \cite{he2016deep, krizhevsky2012imagenet, simonyan2014very, szegedy2015going, Zagoruyko2016WRN}.
Numerous recent work \cite{chen2017sca, Hu_2018, park2018bam, Woo_2018_ECCV,liu2020improving} have proposed using either
channel attention, or spatial attention, or both to 
improve the performance of these neural networks. 
These attention mechanisms have the capabilities of 
improving the feature representations generated by 
standard convolutional layers by explicitly building
dependencies among channels or weighted spatial mask 
for spatial attention. 
The intuition behind learning attention weights is to 
allow the network to have the ability to learn where to attend
and further focus on the target objects.

One of the most prominent methods is the
squeeze-and-excitation networks (SENet) \cite{Hu_2018}. 
Squeeze and Excite (SE) module computes channel attentions and 
provides incremental performance gains at a considerably low cost.
SENet was succeeded by Convolutional Block Attention
Module (CBAM) \cite{Woo_2018_ECCV} and Bottleneck Attention Module
(BAM) \cite{park2018bam}, both of which stressed on providing
robust representative attentions by incorporating spatial attention 
along with channel attention.
They provided substantial performance gains over their 
squeeze-and-excite counterpart at a small computational overhead. 

Different from the aforementioned attention approaches that
require a number of extra learnable parameters, the foundation
backbone of this paper is to investigate the way of building cheap
but effective attentions while maintaining similar or providing 
better performance. 
In particular, we aim to stress on the importance of capturing 
cross-dimension interaction while computing attention weights
to provide rich feature representations.
We take inspiration from the method of computing attention 
in CBAM \cite{Woo_2018_ECCV} which successfully demonstrated
the importance of capturing spatial attention along 
with channel attention. 
In CBAM, the channel attention is computed 
in a similar way as that of SENet \cite{Hu_2018} except for the
usage of global average pooling (GAP) and global max pooling (GMP)
while the spatial attention is generated by simply reducing 
the input to a single channel output to obtain the attention weights.
We observe that the channel attention method within
CBAM \cite{Woo_2018_ECCV} although providing significant 
performance improvements does not account for cross-dimension
interaction which we showcase to have a favorable impact on 
the performance when captured. 
Additionally, CBAM incorporates dimensionality reduction while
computing channel attention which is redundant to capture 
non-linear local dependencies between channels. 

\begin{figure*}[tp]
  \centering
%    \fbox{\rule{0pt}{2in} \rule{.9\linewidth}{0pt}}
  \includegraphics[width=\textwidth]{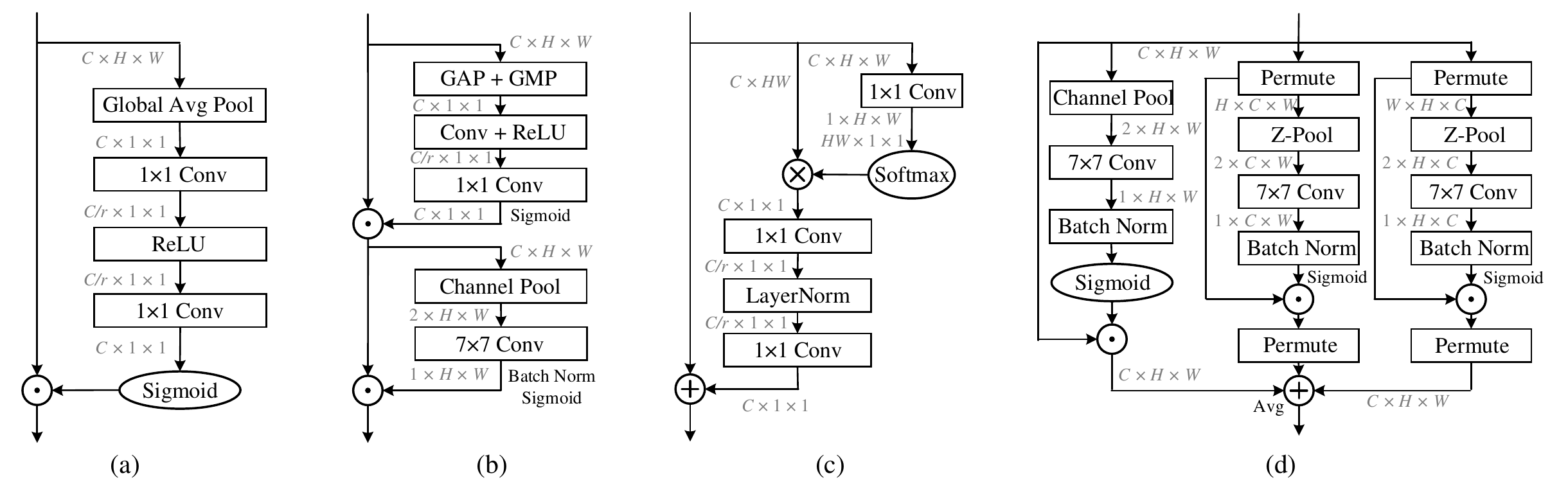} 
  \caption{\textbf{Comparisons with different attention modules:} 
  (a) Squeeze Excitation (SE) Module; (b) Convolutional Block Attention Module (CBAM); (c) Global Context (GC) Module;
  (d) triplet attention (ours). The feature maps are denoted as feature dimensions, \eg$C \times H \times W$ denotes a feature map with channel number $C$, height $H$ and width $W$. $\otimes$ represents matrix multiplication, $\odot$ denotes broadcast element wise multiplication and $\oplus$ denotes broadcast element-wise addition.}
  \label{fig:comps}
\end{figure*}

Based on the above observation, in this paper, we propose
\textit{triplet attention} which accounts for cross-dimension
interaction in an efficient way.
Triplet attention comprises of three branches each responsible for capturing cross-dimension between the spatial dimensions and channel dimension of the input.
Given an input tensor with shape $(C \times H \times W)$, 
each branch is responsible for aggregating 
cross-dimensional interactive features between
either the spatial dimension $H$ or $W$ and 
the channel dimension $C$.
We achieve this by simply permuting the input tensors
in each branch and then passing the tensor
through a \textit{Z-pool}, followed by a convolutional layer 
with kernel size of $k \times k$.
The attention weights are then generated by a sigmoid 
activation layer and then is applied on the permuted 
input tensor before permuting it back into the original 
input shape.

Compared to previous channel attention mechanisms \cite{cao2019gcnet, gao2019global, Hu_2018, park2018bam, Woo_2018_ECCV},
our approach offers two advantages.
First, our method helps in capturing rich discriminative 
feature representations at a negligible computational overhead
which we further empirically verify by visualizing the 
Grad-CAM \cite{selvaraju2017grad} and 
Grad-CAM++ \cite{chattopadhay2018grad} results.
Second, unlike our predecessors, our method stresses
the importance of cross-dimension interaction with 
no dimensionality reduction, thus eliminating indirect 
correspondence between channels and weights.

We showcase this way of computing attention in parallel across
branches while accounting for cross-dimension dependencies 
is extremely effective and cheap in computational terms.
For instance, for ResNet-50 \cite{he2016deep} with 25.557M 
parameters and 4.122 GFLOPs, our proposed plug-in 
triplet attention results in an increase of 
parameters by 4.8K and GFLOPs by 4.7e-2 respectively while 
providing a 2.28$\%$ improvement in Top-1 accuracy. 
We evaluate our method on ImageNet-1k \cite{deng2009imagenet}
classification and object detection on PASCAL VOC \cite{everingham2010pascal} and MS COCO \cite{lin2014microsoft} 
while also providing extensive insight into the effectiveness 
of our method by visualizing the Grad-CAM \cite{selvaraju2017grad} 
and Grad-CAM++ \cite{chattopadhay2018grad} outputs respectively.

\section{Related Work} 

Attention in human perception relates to the process 
of selectively concentrating on parts of the given information 
while ignoring the rest.
This mechanism helps in refining perceived information 
while retaining the context of it. 
Over the last few years, several researched methods have proposed
to efficiently incorporate this attention mechanism 
in deep convolution neural network (CNN) architectures to
improve performance on large-scale vision tasks.
In the following part of this section, we will review
some attention mechanisms that are strongly related to
this work.

Residual Attention Network \cite{wang2017residual} proposes a trunk-and-mask
encoder-decoder style module to generate robust 
three-dimensional attention maps.
Due to the direct generation of 3D attention maps, 
the method is quite computationally complex as compared 
to the recently proposed methods to compute attention. 
This was followed by the introduction of
Squeeze-and-Excitation Networks (SENet) \cite{Hu_2018} which as debated 
by many was the first to successfully implement 
an efficient way of computing channel attention 
while providing significant performance improvements.
The aim of SENet was to model the cross-channel 
relationships in feature maps by learning per-channel
modulation weights. 
Succeeding SENet, Convolutional Block Attention Module
(CBAM) \cite{Woo_2018_ECCV} was proposed, in which they enrich the attention 
maps by adding max pooled features for the channel 
attention along with an added spatial attention component. 
This combination of spatial attention and channel attention 
demonstrated substantial improvement in performance as 
compared to SENet.
More recently, Double Attention Networks (${A}^{2}$-Nets) \cite{chen20182} introduced a novel
relation function for Non-Local (NL) blocks.
NL blocks \cite{wang2018non} were introduced to capture long range dependencies via non-local operations and were designed
to be lightweight and easy to use in any architecture.
Global Second order Pooling Networks (GSoP-Net) \cite{gao2019global} uses 
second-order pooling for richer feature aggregation.
The key idea is to gather important features from 
the entire input space using second order pooling 
and subsequently distributing them to make it easier 
for further layers to recognize and propagate.
Global-Context Networks (GC-Net) \cite{cao2019gcnet} propose a novel NL-block
integrated with a SE block in which they aimed to combine contextual representations with channel weighting more efficiently.
Instead of simple downsampling by GAP as in the case of
SENet \cite{Hu_2018}, GC-Net uses a set of complex permutation-based
operations to reduce the feature maps before passing it 
to the SE block. 

Attention mechanisms have also been successfully used for image segmentation and fine grained image classification. Criss-Cross Networks (CCNet) \cite{huang2019ccnet} and SPNet \cite{hou2020strip} present novel attention blocks to capture rich contextual information using intersecting strips. Xiao \etal \cite{xiao2015application} propose a pipeline integrated with one bottom-up and two top-down attention for fine grained image classification. Cao \etal \cite{cao2015look} introduce the 'Look and Think Twice' mechanism which is based on a computational feedback process inspired from the human visual cortex which helps in capturing visual attention on target objects even in distorted background conditions. 

Most of the above methods have significant shortcomings 
which we address in our method.
Our triplet attention module aims to capture cross-dimension
interaction and thus  is able to provide significant performance
gains at a justified negligible computational overhead 
as compared to the above described methods where 
none of them account for cross-dimension interaction 
while allowing some form of dimensionality reduction 
which is unnecessary to capture cross-channel interaction. 

\section{Proposed Method}

\begin{figure*}
  \centering
%    \fbox{\rule{0pt}{2in} \rule{.9\linewidth}{0pt}}
  \includegraphics[width=0.9\textwidth]{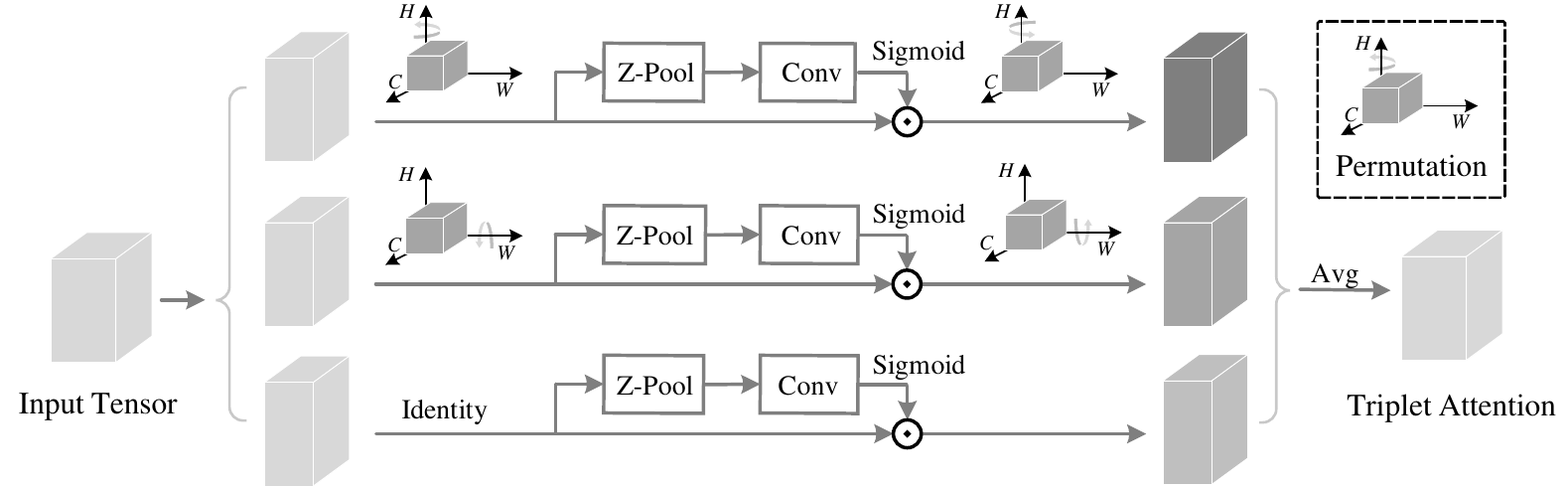} 
  \caption{Illustration of the proposed triplet attention which has
  three branches. The top branch is responsible for computing attention weights across the channel dimension $C$ and the spatial dimension $W$. Similarly, the middle branch is responsible for channel dimension $C$ and spatial dimension $H$. The final branch at the bottom is used to capture spatial dependencies ($H$ and $W$). In the first two branches, we adopt rotation operation to
  build connections between the channel dimension and either one of the spatial dimension. Finally, the weights are aggregated by simple averaging.
  More details can be found in Sec.~\ref{sec:triplet_attention}}
  \label{fig:TripAtt}
\end{figure*}

In this section, we first revisit CBAM \cite{Woo_2018_ECCV} 
and analytically diagnose the efficiency of the shared 
MLP structure within the channel attention module of CBAM. 
Subsequently, we propose our triplet attention module where
we demonstrate the importance of cross-dimension
dependencies and further compare the complexity of our method
with other standard attention mechanisms.
Finally, we conclude by showcasing 
how to adapt triplet attention into standard deep CNN 
architectures for different challenging tasks in the 
domain of computer vision. 

\subsection{Revisiting Channel Attention in CBAM}

We first revisit the channel attention module used in CBAM \cite{Woo_2018_ECCV} in this subsection.
Let $\chi \in \mathbb{R}^{C \times H \times W} $ 
be the output of a convolutional layer and the subsequent input to 
the channel attention module of CBAM 
where $C$, $H$ and $W$ represent the channels of the tenor 
or the number of filters, height, and width of 
the spatial feature maps, respectively.
The channel attention in CBAM can be represented 
by the following equation: 
\begin{equation} \label{eq:1}
\omega = \sigma(f_{(\textbf{W}_0, \textbf{W}_1)}(g(\chi)) + f_{(\textbf{W}_0, \textbf{W}_1)}(\delta(\chi)))
\end{equation}
where $\omega \in \mathbb{R}^{C \times 1 \times 1}$ 
represent the learnt channel attention weights 
which are then applied to the input $\chi$,
$g(\chi)$ is the global average pooling (GAP) function
as formulated as follows:
\begin{equation}
    g(\chi) = \frac{1}{W \times H} \sum_{i=1}^{H}\sum_{j=1}^{W} \chi_{i,j}
\end{equation}
and 
$\delta(\chi)$ represents the global max pooling (GMP) function
written as:
\begin{equation}
    \delta(\chi) = \max_{H,W}(\chi)
\end{equation}
The above two pooling functions make up the two methods of 
spatial feature aggregation in CBAM. 
Symbol $\sigma$ represents the sigmoid activation function.
Functions $f_{(\mathbf{W}_0, \mathbf{W}_1)}(g(\chi))$ and 
$f_{(\mathbf{W}_0, \mathbf{W}_1)}(\delta(\chi))$ are two transformations.
Thus, after expanding $f_{(\mathbf{W}_0, \mathbf{W}_1)}(g(\chi))$ and 
$f_{(\mathbf{W}_0, \mathbf{W}_1)}(\delta(\chi))$, we have the following form of $\omega$:  
\begin{equation}
    \omega = \sigma(\mathbf{W}_{1}\mbox{ReLU}(\mathbf{W}_{0}g(\chi)) + \mathbf{W}_{1}\mbox{ReLU}(\mathbf{W}_{0}\delta(\chi)))  \label{eq:2}
\end{equation}
where ReLU represents the Rectified Linear Unit and 
$\mathbf{W}_{0}$ and $\mathbf{W}_{1}$ are weight matrices,
the size of which are defined to be $C \times \frac{C}{r}$ and $\frac{C}{r} \times C$, respectively. 
Here, $r$ represents the reduction ratio in the bottleneck of the MLP network
which is responsible for dimensionality reduction. 
Larger $r$ results in lower computational complexity and vice versa. 
To note, the weights of the MLP: $\mathbf{W}_{0}$ and $\mathbf{W}_{1}$ are shared in CBAM 
for both the inputs: $g(\chi)$ and $\delta(\chi)$. 
In Eq. \eqref{eq:2}, the channel descriptors are projected into a lower dimensional space
and then maps them back which causes loss in inter-channel relation 
because of the indirect weight-channel correspondence. 

\subsection{Triplet Attention} \label{sec:triplet_attention}

As demonstrated in Sec.~\ref{sec:introduction}, the goal of
this paper is to investigate how to model cheap but effective
channel attention while not involving any dimensionality reduction. 
In this subsection, unlike CBAM \cite{Woo_2018_ECCV} and SENet \cite{Hu_2018}, which require a certain
number of learnable parameters to build inter-dependencies
among channels, we present an almost parameter-free attention
mechanism to model channel attention and spatial attention, namely
triplet attention.

\paragraph{Overview:}
The diagram of the proposed triplet attention can be found in
Fig. \ref{fig:TripAtt}. 
As the name implies, triplet attention is made up of three 
parallel branches, two of which are responsible for 
capturing cross-dimension interaction between the channel 
dimension $C$ and either the spatial dimension $H$ or $W$.
The remaining final branch is similar to
CBAM \cite{Woo_2018_ECCV}, used to build spatial attention.
The outputs from all three branches are aggregated using 
simple averaging.
In the following, before specifically describing the proposed
triplet attention, we first introduce the intuition of
building cross-dimension interaction.
% \hou{Here needs some improvement. I think we need to briefly
% describe what looks like in each branch.}

\paragraph{Cross-Dimension Interaction:} Traditional ways 
of computing channel attention involve computing a singular 
weight, often a scalar for each channel in the input tensor 
and then scaling these feature maps uniformly using the 
singular weight. 
Though this process of computing channel attention has been 
proven to be extremely lightweight and quite successful, 
there is a significant missing piece in considering this method.
Usually, to compute these singular weights for channels, the input 
tensor is spatially decomposed to one pixel per channel by 
performing global average pooling.
This results in a major loss of spatial information and thus 
the inter-dependence between the channel dimension and the spatial
dimension is absent when computing attention on these single 
pixel channels.
CBAM \cite{Woo_2018_ECCV} introduced spatial attention 
as a complementary module to the channel attention.
In simple terms, the spatial attention tells '\textit{where in the channel 
to focus}' and the channel attention tells '\textit{what channel to 
focus on}'.
However, the shortcoming in this process is that the channel
attention and spatial attention are segregated and computed
independent of each other. Thus, any relationship
between the two is not considered. 
Motivated by the way of building spatial attention, we present
the concept of \textit{cross dimension interaction}, which
addresses this shortcoming by capturing the interaction 
between the spatial dimensions and the channel dimension
of the input tensor.
We introduce cross-dimension interaction in triplet attention by 
dedicating three branches to capture dependencies between the ($C$, $H$), ($C$, $W$) and ($H$, $W$) dimensions of the input tensor respectively.

\paragraph{Z-pool:} The Z-pool layer here is responsible for reducing the zeroth
dimension of the tensor to two by concatenating
the average pooled and max pooled features across that dimension.
This enables the layer to preserve a rich representation of the actual tensor 
while simultaneously shrinking its depth to make further
computation lightweight.
Mathematically, it can be represented by the following equation:
\begin{equation}
    Z\text{-pool}(\chi) = [\text{MaxPool}_{0d}(\chi), \text{AvgPool}_{0d}(\chi)],
\end{equation}
where $0d$ is the 0th-dimension across which the max and average pooling operations take place.
For instance, the Z-Pool of a tensor of shape $(C \times H \times W)$ results in
a tensor of shape $(2 \times H \times W)$.

\paragraph{Triplet Attention:} Given the above defined operations,
we define \textit{triplet attention} as a three branched module
which takes in an input tensor and outputs a refined tensor of 
the same shape. 
Given an input tensor $\chi \in \mathbb{R}^{C \times H \times W}$,
we first pass it to each of the three branches in the proposed
triplet attention module.
In the first branch, we build interactions between the height 
dimension and the channel dimension.
To achieve so, the input $\chi$ is rotated $90^{\circ}$
anti-clockwise along the $H$ axis.
This rotated tensor denoted as $\thickhat{\chi_{1}}$ is of 
the shape $(W \times H \times C)$. 
$\thickhat{\chi_{1}}$ is then passed through Z-pool and
is subsequently reduced to $\thickhat{\chi_{1}^{\ast}}$ 
which is of shape $(2 \times H \times C)$.
$\thickhat{\chi_{1}^{\ast}}$ is then passed through 
a standard convolutional layer of kernel size $k \times k$
followed by a batch normalization layer, which provides
the intermediate output of dimensions $(1 \times H \times C)$. 
The resultant attention weights are then generated 
by passing the tensor through a sigmoid activation
layer ($\sigma$).
The attention weights generated are subsequently applied to
$\thickhat{\chi_{1}}$ and then rotated 90$^{\circ}$ clockwise 
along the $H$ axis to retain the original input shape of $\chi$.

Similarly, in the second branch, we rotate $\chi$ 90$^{\circ}$ anti-clockwise along the $W$ axis.
The rotated tensor $\thickhat{\chi_{2}}$ can be represented 
with dimension of $(H \times C \times W)$ and is passed 
through a Z-pool layer.
Thus, the tensor is reduced to $\thickhat{\chi_{2}^{\ast}}$ 
of the shape $(2 \times C \times W)$.
$\thickhat{\chi_{2}^{\ast}}$ is passed through a standard
convolutional layer defined by kernel size $k \times k$ 
followed by a batch normalization layer which outputs
a tensor of the shape $(1 \times C \times W)$. 
The attention weights are then obtained by passing 
this tensor through a sigmoid activation layer ($\sigma$) 
which are then simply applied on  $\thickhat{\chi_{2}}$ 
and the output is subsequently rotated 90$^{\circ}$ clockwise 
along the $W$ axis to retain the same shape as input $\chi$. 

For the final branch, the channels of the input tensor $\chi$ are reduced to two by Z-pool. This reduced tensor $\thickhat{\chi_{3}}$ of shape $(2 \times H \times W)$ is then passed through a standard convolution layer defined by kernel size $k$ followed by a batch normalization layer. The output is passed through sigmoid activation layer ($\sigma$) to generate the attention weights of shape $(1 \times H \times W)$ which are then applied to the input $\chi$. The refined tensors of shape $(C \times H \times W)$ generated by each of the three branches are then aggregated by simple averaging. 

Summarizing, the process to obtain the refined attention-applied tensor $y$ from triplet attention for an input tensor $\chi \in \mathbb{R}^{C \times H \times W}$ can be represented by the following equation: 
\begin{equation}
    y = \frac{1}{3}(\overline{\thickhat{\chi_{1}}\sigma(\psi_{1}(\thickhat{\chi_{1}^{\ast}}))} + \overline{\thickhat{\chi_{2}}\sigma(\psi_{2}(\thickhat{\chi_{2}^{\ast}}))} + \chi \sigma(\psi_{3}(\thickhat{\chi_{3}}))),
    \label{eq:3}
\end{equation}
where $\sigma$ represents the sigmoid activation function; $\psi_{1}$, $\psi_{2}$ and $\psi_{3}$ represent the standard two-dimensional convolutional layers defined by kernel size $k$
in the three branches of triplet attention. 
Simplifying Eq.\eqref{eq:3}, $y$ becomes:
\begin{equation}
    y = \frac{1}{3}(\overline{\thickhat{\chi_{1}}\omega_{1}} + \overline{\thickhat{\chi_{2}}\omega_{2}} + \chi\omega_{3}) = \frac{1}{3}(\overline{y_{1}} + \overline{y_{2}} + y_{3}),
    \label{eq:4}
\end{equation}
where $\omega_{1}$, $\omega_{2}$ and $\omega_{3}$ are the three cross-dimensional attention weights computed in triplet attention. The $\overline{y_{1}}$ and $\overline{y_{2}}$ in Eq. \eqref{eq:4} represents the 90$^{\circ}$ clockwise rotation to retain the original input shape of $(C \times H \times W)$.

\begin{table}
\renewcommand{\arraystretch}{1.2}
\begin{center}
\resizebox{\columnwidth}{!}{
\begin{tabular}{|l|c |c|}
\hline
Attention Mechanism & Parameters & Overhead (ResNet-50) \\
\hline\hline
SE \cite{Hu_2018} & $2C^{2}/r$ & 2.514M\\
CBAM \cite{Woo_2018_ECCV} & $2C^{2}/r + 2k^{2}$ & 2.532M\\
BAM \cite{park2018bam} & $C/r(3C + 2k^{2}C/r + 1)$ & 0.358M\\
GC \cite{cao2019gcnet} & $2C^{2}/r + C$ & 2.548M\\
Triplet Attention & $6k^{2}$ & 0.0048M\\
\hline
\end{tabular}}
\end{center}
\caption{Comparisons of various attention modules based on their parameter complexity and overhead using ResNet-50 backbone.}
\label{tab:complexity}
\end{table}

\paragraph{Complexity Analysis:}
In Tab. \ref{tab:complexity}, we empirically verify the parameter efficiency of triplet attention as compared to other standard attention mechanisms. $C$ represents the number of input channels to the layer, $r$ represents the reduction ratio used in the bottleneck of the MLP while computing the channel attention and the kernel size used for 2D convolution is represented by $k$; $k \lll C$. We show that the parameter overhead brought along by different attention layers is much higher as compared to our method. We calculate the overhead on a ResNet-50 \cite{he2016deep} by adding the attention layers in each block while fixing $r$ to be 16. $k$ was fixed at 7 for CBAM \cite{Woo_2018_ECCV} and triplet attention while for BAM \cite{park2018bam} $k$ was set to be 3. The reason for the lower overhead cost for BAM as compared to CBAM, GC \cite{cao2019gcnet} and SE \cite{Hu_2018} is because unlike the latter mentioned attention layers being used in every block, BAM was used only three times across the architecture in total according to the default setting for BAM.

\begin{table*}
\small
\setlength\tabcolsep{4mm}
\renewcommand{\arraystretch}{1.2}
\begin{center}
\begin{tabular}{|l|c|c|c|c|c|}
\hline
Method & Backbone & Parameters & FLOPs & Top-1 (\%) & Top-5 (\%) \\
\hline\hline
\multirow{3}{*}{ResNet \cite{he2016deep}} & ResNet-18 & \textbf{11.69M} & \textbf{1.82G} & 30.20 & 10.90\\
 & ResNet-50 & \textbf{25.56M} & \textbf{4.12G} & 24.56 & 7.50\\
 & ResNet-101 & \textbf{44.46M} & \textbf{7.85G} & 22.63 & 6.44\\ \hline
SENet \cite{Hu_2018} & \multirow{4}{*}{ResNet-18} & 11.78M & 1.82G & 29.41 & 10.22\\
BAM \cite{park2018bam} & & 11.71M & 1.83G & \textbf{28.88} & \textbf{10.01} \\
CBAM \cite{Woo_2018_ECCV} & & 11.78M & 1.82G & 29.27 & 10.09\\
Triplet Attention (Ours) & & \textbf{11.69M} & 1.83G & \textbf{28.91} & \textbf{10.01}\\
\hline
SENet \cite{Hu_2018} & \multirow{9}{*}{ResNet-50} & 28.07M & 4.13G & 23.14 & 6.70\\
BAM \cite{park2018bam} & & 25.92M & 4.21G & 24.02 & 7.18 \\
CBAM \cite{Woo_2018_ECCV} & & 28.09M & 4.13G & 22.66 & 6.31\\
GSoP-Net1 \cite{gao2019global} & & 28.29M & 6.41G & \textbf{22.02} & \textbf{5.88}\\
${A}^{2}$-Nets \cite{chen20182} & & 33.00M & 6.50G & 23.00 & 6.50\\
GCNet \cite{cao2019gcnet} & & 28.10M & 4.13G & 22.30 & 6.34\\
GALA \cite{linsley2018learning} &  & 29.40M & - & 22.73 & 6.35 \\ 
ABN \cite{Fukui_2019_CVPR} & & 43.59M & 7.66G & 23.10 & - \\
SRM \cite{lee2019srm} & & 25.62M & 4.12G & 22.87 & 6.49 \\
Triplet Attention (Ours) & & \textbf{25.56M} & 4.17G & \textbf{22.52} & \textbf{6.32}\\
\hline
SENet \cite{Hu_2018} & \multirow{5}{*}{ResNet-101} & 49.29M & 7.86G & 22.38 & 6.07\\
BAM \cite{park2018bam} & & 44.91M & 7.93G & 22.44 & 6.29 \\
CBAM \cite{Woo_2018_ECCV} & & 49.33M & 7.86G & \textbf{21.51} & \textbf{5.69}\\
SRM \cite{lee2019srm} & & 44.68M & 7.85G & 21.53 & 5.80\\
Triplet Attention (Ours) & & \textbf{44.56M} & 7.95G & \textbf{21.97} & \textbf{6.15}\\
\hline \hline
MobileNetV2 \cite{sandler2018mobilenetv2} & \multirow{4}{*}{MobileNetV2} & \textbf{3.51M} & \textbf{0.32G} & 28.36 & 9.80\\
SENet \cite{Hu_2018} & & 3.53M & 0.32G & 27.58 & 9.33\\
CBAM \cite{Woo_2018_ECCV} & & 3.54M & 0.32G & 30.07 & 10.67\\
Triplet Attention (Ours) & & \textbf{3.51M} & 0.32G & \textbf{27.38} & \textbf{9.23}\\
\hline
\end{tabular}
\end{center}
\caption{Single-crop error rate ($\%$) on the ImageNet validation set
and complexity comparisons in terms of network parameters (in millions) and floating point operations per second (FLOPs). Other than reporting results on heavy-weight ResNets, we
also show results based on light-weight mobile networks. 
With a negligible increase of learnable parameters, our approach works much better than the
baselines and is also comparable to the state-of-the-art
methods that need large additional parameters and
computations, like GSoP-Net1 \cite{gao2019global}.
}
\label{tab:imagenet}
\end{table*}

\section{Experiments}

In this section, we provide the details for experiments 
and results that demonstrate the performance and efficiency 
of triplet attention, and compare it with previously
proposed attention mechanisms on several challenging
computer vision tasks like ImageNet-1k \cite{deng2009imagenet} classification and object detection on PASCAL VOC \cite{everingham2010pascal} and MS COCO \cite{lin2014microsoft} datasets using standard network architectures like ResNet-50 \cite{he2016deep} and 
MobileNetV2 \cite{sandler2018mobilenetv2}.
To further validate our results, we provide the Grad-CAM \cite{selvaraju2017grad} and Grad-CAM++ \cite{chattopadhay2018grad} results for sample images 
to showcase the ability of triplet attention to capture
more deterministic feature-rich representations. 

All ImageNet models were trained using 8 Nvidia Tesla V100 GPUs, and all object detection models were trained with 4 Nvidia Tesla P100 GPUs. We did not observe any substantial difference in total wall time between the baseline models and those augmented with triplet attention.

\subsection{ImageNet} \label{ImageNet}

To train our ResNet \cite{he2016deep} based models, 
we add triplet attention layers at the end of each 
bottleneck block. 
We follow the exact training configuration as \cite{he2016deep, Hu_2018} for consistent and fair comparison with other methods. 
Similarly, we follow the approach of \cite{sandler2018mobilenetv2} to
train our MobileNetV2-based architecture.

Our results for the validated architectures are
shown in Tab. \ref{tab:imagenet}.
Triplet attention is able to match or outperform other 
similar techniques, while simultaneously introducing
the fewest number of additional model parameters. 
\begin{table*}[t]
\setlength\tabcolsep{2.8mm}
\small
\renewcommand{\arraystretch}{1.2}
\begin{center}
\begin{tabular}{|l|c|c|c|c|c|c|c|c|}
\hline
Backbone & Detectors & Parameters & AP & AP$_{50}$ & AP$_{75}$ & AP$_{S}$ & AP$_{M}$ & AP$_{L}$\\
\hline\hline
ResNet-50 \cite{he2016deep} & \multirow{5}{*}{Faster R-CNN \cite{ren2015faster}} & \textbf{41.7M} & 36.4 & 58.4 & 39.1 & 21.5 & 40.0 & 46.6\\
ResNet-101 \cite{he2016deep} & & 60.6M & 38.5 & 60.3 & 41.6 & 22.3 & \textbf{43.0} & 49.8 \\
SENet-50 \cite{Hu_2018} & & 44.2M & 37.7 & 60.1 & 40.9 & 22.9 & 41.9 & 48.2 \\
ResNet-50 + CBAM \cite{Woo_2018_ECCV} & & 44.2M & \textbf{39.3} & \textbf{60.8} & \textbf{42.8} & \textbf{24.1} & \textbf{43.0} & 49.8\\
ResNet-50 + Triplet Attention (Ours) & & \textbf{41.7M} & \textbf{39.3} & \textbf{60.8} & 42.7 & 23.4 & 42.8 & \textbf{50.3}\\
\hline
ResNet-50 \cite{he2016deep} & \multirow{4}{*}{RetinaNet \cite{ren2015faster}} & \textbf{38.0M} & 35.6 & 55.5 & 38.3 & 20.0 & 39.6 & 46.8\\
SENet-50 \cite{Hu_2018} & & 40.5M & 37.1 & 57.2 & 39.9 & 21.2 & 40.7 & 49.3 \\
ResNet-50 + CBAM \cite{Woo_2018_ECCV} & & 40.5M & \textbf{38.0} & \textbf{57.7} & \textbf{40.6} & \textbf{22.1} & \textbf{41.9} & \textbf{50.2}\\
ResNet-50 + Triplet Attention (Ours) & & \textbf{38.0M} & 37.6 & 57.3 & 40.0 & 21.7 & 41.1 & 49.7 \\
\hline
ResNet-50 \cite{he2016deep} & \multirow{5}{*}{Mask RCNN \cite{he2017mask}} & \textbf{44.3M} & 37.3 & 59.0 & 40.2 & 21.9 & 40.9 & 48.1\\
SENet-50 \cite{Hu_2018} & & 46.8M & 38.7 & 60.9 & 42.1 & 23.4 & 42.7 & 50.0 \\
ResNet-50 + 1 NL block \cite{wang2018non} & & 46.5M & 38.0 & 59.8 & 41.0 & - & - & - \\
GCNet \cite{gao2019global} & & 46.9M & 39.4 & \textbf{61.6} & 42.4 & - & - & - \\
ResNet-50 + Triplet Attention (Ours) & & \textbf{44.3M} & \textbf{39.8} & \textbf{61.6} & \textbf{42.8}  & \textbf{24.3} & \textbf{42.9} & \textbf{51.3} \\
\hline
\end{tabular}
\end{center}
\caption{\textbf{Object detection mAP($\%$) on the MS COCO validation set}. Triplet Attention results in higher performance gain with minimal computational overhead.}
\label{tab:coco}
\end{table*}

A ResNet50-based model augmented with triplet attention 
achieves a 2.04\% improvement in top-1 error rate on 
ImageNet while only increasing the number of parameters
by approximately 0.02\% and increasing the FLOPs by $\approx$1\%.
The only comparable model that outperforms triplet attention
is GSoP-Net, which provides a 0.5\% gain over triplet
attention at the cost of 10.7\% more parameters and 53.6\% 
more FLOPs.

We observe similar trend in performance in the smaller ResNet-18 model where triplet attention provides a 0.5\% improvement in top-1 error rate while only increasing the parametric complexity by 0.02\%.

For ResNet-101 based models, triplet attention outperforms both vanilla
and SENet variants by 0.66\% and 0.41\%, respectively. While SRM \cite{lee2019srm} and CBAM
were able to obtain marginally better results than triplet attention, our approach is still 
the lightest in terms of parameters. 

With MobileNetV2, triplet attention provides a 0.98\% 
improvement in top-1 error rate on ImageNet while only 
increasing parameters by approximately 0.03\%. 
We also observed that CBAM hurts model performance in case
of a MobileNetV2 where it drops accuracy by 1.71\%.  
The experimental results demonstrate that the proposed triplet
attention works well for both heavy and light-weight models
with a negligible increase in parameters and computations.
In the following subsection and supplementary materials,
we will show the effectiveness of our triplet attention module
when applied to other vision tasks, like object detection,
instance segmentation, and human key-point detection.

\subsection{PASCAL VOC} \label{pascal}

\begin{table}[H]
\renewcommand{\arraystretch}{1.2}
\begin{center}
\resizebox{\columnwidth}{!}{
\begin{tabular}{|l|c|c|c|c|}
\hline
Method & Detector & $AP$ & $AP_{50}$ & $AP_{75}$ \\
\hline\hline
ResNet-50 \cite{he2016deep} & \multirow{3}{*}{Faster R-CNN \cite{ren2015faster}} & 46.956 & 77.521 & 48.903\\
ResNet-50 + CBAM \cite{Woo_2018_ECCV} & & 51.398 & 80.409 & 54.919\\
ResNet-50 + TA (Ours) & & \textbf{53.919} & \textbf{80.932} & \textbf{58.810}\\
\hline
\end{tabular}}
\end{center}
\caption{\textbf{Object detection mAP($\%$) on the PASCAL VOC 2012 test set}. Triplet attention results in providing significant improvement in performance with negligible overhead as compared to it's counterparts. TA represents Triplet Attention.}
\label{tab:VOC}
\end{table}

For object detection, we utilize our pre-trained ResNet-50
model described in Sec. \ref{ImageNet} in conjunction with 
Faster R-CNN \cite{ren2015faster} with 
FPN \cite{lin2017feature} on the Pascal VOC dataset \cite{everingham2010pascal}. 
We adopt default training configuration for the
detectron2 toolkit \cite{wu2019detectron2} to 
train a baseline ResNet-50 \cite{he2016deep} and
ResNet-50 with CBAM \cite{Woo_2018_ECCV}.
For all models, we train on the 2007 and 2012 versions
of the training set and validate on the 2007 validation set
as described in \cite{everingham2010pascal}.

The results can be found in Tab.~\ref{tab:VOC}. 
When compared to the baseline model and its corresponding CBAM variant,
our triplet attention module is able to produce a distinct 
improvement in AP score, beating the baseline ResNet50 
by 6.9\%, and CBAM by 2.6\% 
while having a backbone that consumes fewer FLOPs 
and parameters.

\subsection{MS COCO} \label{coco}

As in Sec. \ref{pascal}, using the ImageNet models augmented with triplet attention as backbones, we train Faster-RCNN \cite{ren2015faster}, Mask-RCNN \cite{he2017mask}, and RetinaNet \cite{lin2017focal} models to apply our attention module to object detection tasks on the COCO dataset \cite{lin2014microsoft}.
We use the training procedure described in \cite{lin2017focal, ren2015faster},implemented in the mmdetection framework\cite{chen2019mmdetection}, to ensure a fair test.
Our results for the COCO dataset results are summarized in Tab. \ref{tab:coco}.
We observe that triplet attention outperforms most of the similar layers, achieving a higher AP score in multiple categories.
Across all architectures, adding a triplet attention module improves the AP score by over 2 points in AP over the baseline model while using the same ImageNet backbone described in Sec. \ref{ImageNet} that adds a negligible computational overhead.
The improvement in performance observed in the experiments showcase the benefit of our cross-dimension interaction strategy in triplet attention.

\subsection{Ablation Study on Branches}

\begin{table}[ht]
\renewcommand{\arraystretch}{1.2}
\begin{center}
\resizebox{\columnwidth}{!}{
\begin{tabular}{|l|c|c|c|}
\hline
Model & Parameters & FLOPs & Top-1 Accuracy (\%) \\
\hline\hline
ResNet-32 \cite{he2016deep} & 0.464M & 3.404G & 93.12\\
ResNet-32 + TA (channel off) & 0.466M & 3.437G & 93.27\\
ResNet-32 + TA (spatial off) & 0.467M & 3.415G & 93.29\\
ResNet-32 + TA (full) & 0.469M & 3.448G & \textbf{93.56}\\
\hline
VGG-16 + BN \cite{simonyan2014very} & 15.254M & 0.315G & 93.25\\
VGG-16 + BN + TA (channel off) & 15.255M & 0.315G & 93.59\\
VGG-16 + BN + TA (spatial off) & 15.256M & 0.32G & 93.15\\
VGG-16 + BN + TA (full) & 15.257M & 0.32G & \textbf{93.78}\\
\hline
MobileNet-v2 \cite{sandler2018mobilenetv2} & 2.297M & 0.095G & 93.11\\
MobileNet-v2 + TA (channel off) & 2.302M & 0.096G & 92.94\\
MobileNet-v2 + TA (spatial off) & 2.308M & 0.12G & 93.22\\
MobileNet-v2 + TA (full) & 2.313M & 0.122G & \textbf{93.51}\\
\hline
\end{tabular}}
\end{center}
\caption{Effect of different branches in triplet attention on performance in CIFAR-10 classification.}
\label{tab:branch}
\end{table}

We further validate the importance of cross-dimension interaction by conducting ablation experiments to observe the impact of the branches in the triplet attention module. In Tab. \ref{tab:branch}, \textit{spatial off} indicates that the third branch, where the input tensor is not permuted, is switched off, and \textit{channel off} indicates that the two branches, which involve permutations of the input tensor, are switched off. As shown, the results support our intuition with triplet attention having all three branches switched on, denoted as \textit{full}, to be performing consistently better than the vanilla version and its two counterparts. 

\subsection{Grad-CAM Visualization}
\begin{figure}[ht]
  \centering
%    \fbox{\rule{0pt}{2in} \rule{.9\linewidth}{0pt}}
  \includegraphics[width=0.45\textwidth]{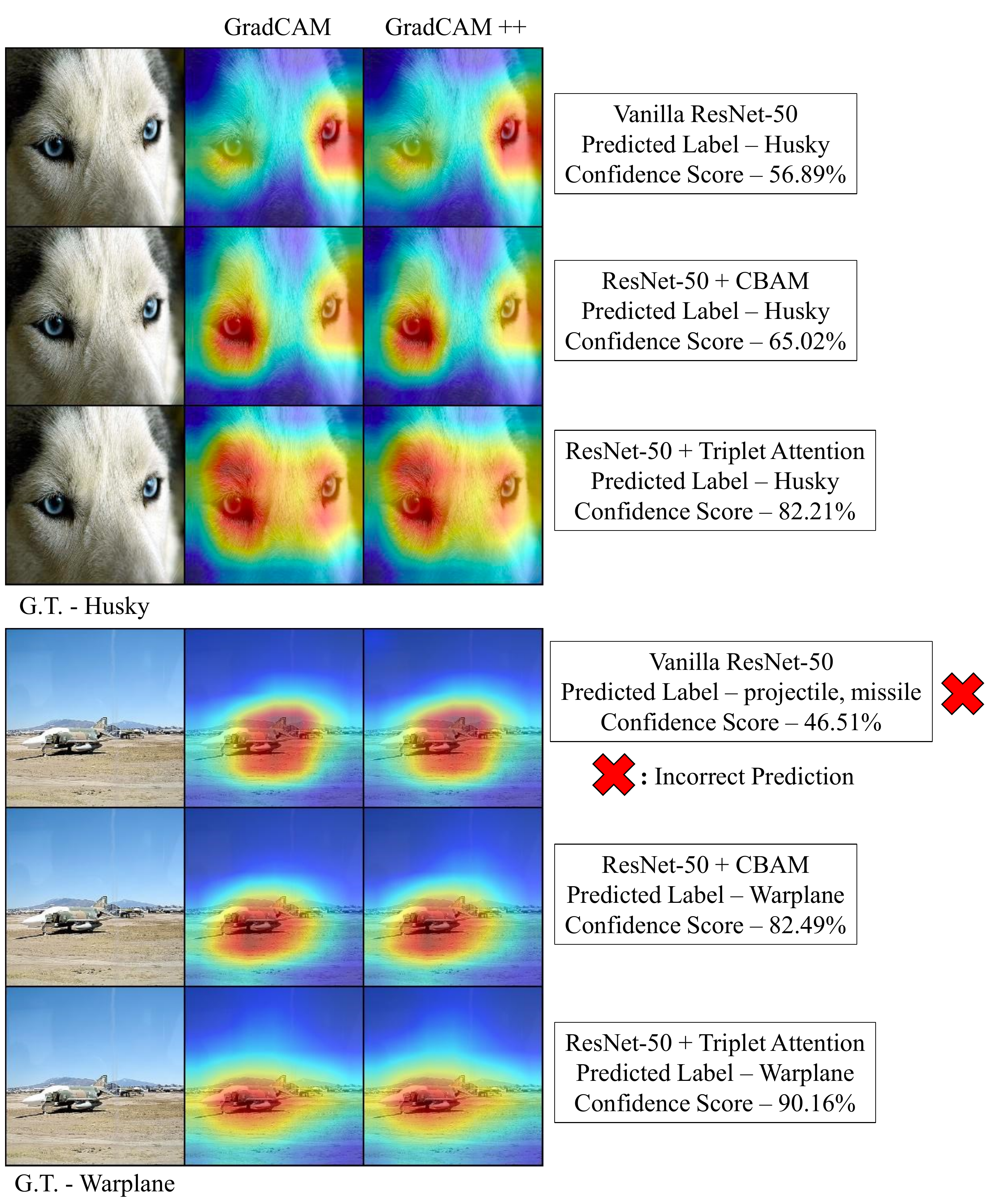} 
  \caption{\textbf{Visualization of Grad-CAM and Grad-CAM++ results.} The results were obtained for two random samples from the ImageNet validation set and were compared for a baseline ResNet-50, ResNet-50 + CBAM and a ResNet-50 + triplet attention. Ground truth (G.T) labels for the images are provided below the original samples and the networks prediction and confidence scores are provided in the corresponding boxes.}
  \label{fig:gradcam}
\end{figure}
We hypothesize that the cross-dimensional interaction provided by triplet attention helps the network learn more meaningful internal representations of the image. To validate this claim, we provide sample visualizations from the Grad-CAM \cite{selvaraju2017grad} and Grad-CAM++ \cite{chattopadhay2018grad} techniques, which visualize the gradients of the top-class prediction with respect to the input image as a colored overlay. 
As shown in Fig.~\ref{fig:gradcam}, triplet attention is able to capture tighter and more relevant bounds on images from the ImageNet dataset \cite{deng2009imagenet}.
In certain cases, when using triplet attention,
a ResNet50 is able to identify classes
that the baseline model fails at predicting correctly. More Grad-CAM based results are presented in the supplementary section.

The visualizations support our understanding of the intrinsic capability of triplet attention to capture richer and more discriminative contextual information for a particular target class. This property of triplet attention is extremely favorable and helpful in improving the performance of deep neural network architectures as compared to their baseline counterparts.

\section{Conclusion}

In this work, we propose a novel attention layer, triplet attention, which captures the importance of features across dimensions in a tensor. Triplet attention uses an efficient attention computation method that does not have any information bottlenecks. Our experiments demonstrate that triplet attention improves the baseline performance of architectures like ResNet and MobileNet on tasks like image classification on ImageNet and object detection on MS COCO, while only introducing a minimal computational overhead.

We expect that other novel and robust techniques of capturing cross-dimension dependencies when computing attention may improve upon our results while reducing cost. In the future, we plan to investigate the effects of adding triplet attention to more sophisticated architectures like EfficientNets \cite{tan2019efficientnet} and extend our intuition in the domain of 3D vision.

\section{Acknowledgements}

The authors would like to offer sincere gratitude to everyone who supported during the timeline of this project including Himanshu Arora from Montreal Institute for Learning Algorithms (MILA), Jaegul Choo and Sanghun Jung from Korea Advanced Institute of Science and Technology (KAIST). This work utilizes resources supported by the National Science Foundation’s Major Research Instrumentation program \cite{kindratenko2020hal}, grant \#1725729, as well as the University of Illinois at Urbana-Champaign.

{\small
\bibliographystyle{ieee_fullname}
\bibliography{egbib}
}

\clearpage
\appendix
\setcounter{equation}{0}
\setcounter{figure}{0}
\setcounter{table}{0}
\setcounter{page}{1}
\setcounter{section}{0}
\makeatletter
\renewcommand{\theequation}{S\arabic{equation}}
\renewcommand{\thefigure}{S\arabic{figure}}

\section{Supplementary Experiments}

In this section, we provide results for additional experiments that we ran to evaluate the performance of triplet attention on other vision tasks adjacent to the main focus on image classification and object detection in the paper.

In particular, we expand our Mask RCNN model to use a keypoint detection head, as specified in \cite{he2017mask}, and evaluate the existing Mask-RCNN model on the COCO instance segmentation task. We also observe the effect of kernel size $k$ in the convolution operations within the triplet attention module added to different standard architectures.

In addition, we provide more GradCAM \cite{selvaraju2017grad} and GradCAM++ \cite{chattopadhay2018grad} visualizations, and observe some interesting patterns in the resulting heatmaps, which we discuss further in Sec. \ref{grad1}.

\section{Effect of kernel size $k$}

\begin{table}[H]
\renewcommand{\arraystretch}{1.2}
\begin{center}
\resizebox{\columnwidth}{!}{
\begin{tabular}{|l|c|c|c|c|c|}
\hline
Architecture & Dataset & $k$ & Param. & FLOPs & Top-1 ($\%$)\\
\hline\hline
\multirow{3}{*}{ResNet-20 \cite{he2016deep}} & \multirow{3}{*}{CIFAR-10} & 3 & \textbf{0.270M} & \textbf{2.011G} & 92.66\\
 & & 5 & 0.271M & 2.019G & 92.71 \\
 & & 7 & 0.272M & 2.032G & \textbf{92.91} \\
 \hline
\multirow{3}{*}{VGG-16 + BN \cite{simonyan2014very}} &  \multirow{3}{*}{CIFAR-10} & 3 & \textbf{15.254M} & \textbf{0.316G} & 91.73 \\
 & & 5 & 15.255M & 0.317G & 92.05 \\
 & & 7 & 15.256M & 0.32G & \textbf{92.33} \\
\hline
\multirow{2}{*}{ResNet-18 \cite{he2016deep}} &  \multirow{2}{*}{ImageNet } & 3 & \textbf{11.69M} & \textbf{1.823G} & 70.33\\
 & & 7 & 11.69M & 1.825G & \textbf{71.09} \\
\hline
\multirow{2}{*}{ResNet-50 \cite{he2016deep}} &  \multirow{2}{*}{ImageNet } & 3 & \textbf{25.558M} & \textbf{4.131G} & 76.12\\
 & & 7 & 25.562M & 4.169G & \textbf{77.48} \\
\hline
\multirow{2}{*}{MobileNetV2 \cite{sandler2018mobilenetv2}} &  \multirow{2}{*}{ImageNet} & 3 & \textbf{3.506M} & \textbf{0.322G} & \textbf{72.62} \\
 & & 7 & 3.51M & 0.327G & 71.99 \\
 \hline
\end{tabular}}
\end{center}
\caption{Effect of kernel size $k$ for triplet attention in standard CNN architectures on CIFAR-10  \cite{krizhevsky2009learning} and ImageNet \cite{deng2009imagenet}. We observe a general trend
of improvement in performance with increasing kernel size aside from MobileNetV2.}
\label{tab:kernel}
\end{table}

We do baseline experiments to compare the effect of
using different kernel sizes $k$ in triplet attention and 
show our results in Tab.~\ref{tab:kernel}.
We conduct experiments on both CIFAR-10 and ImageNet with
different network architectures to demonstrate the flexibility
of the proposed triplet attention.
From Tab.~\ref{tab:kernel}, we observe a general trend
of improvement in performance with increasing kernel size.
When deployed in lighter-weight models, like MobileNetV2 \cite{sandler2018mobilenetv2}, we observed a smaller kernel
to outperform its larger kernel counterpart and thus
overall have less complexity overhead. 

\begin{figure*}
  \centering
%    \fbox{\rule{0pt}{2in} \rule{.9\linewidth}{0pt}}
  \includegraphics[width=\textwidth,height= 0.9\textheight]{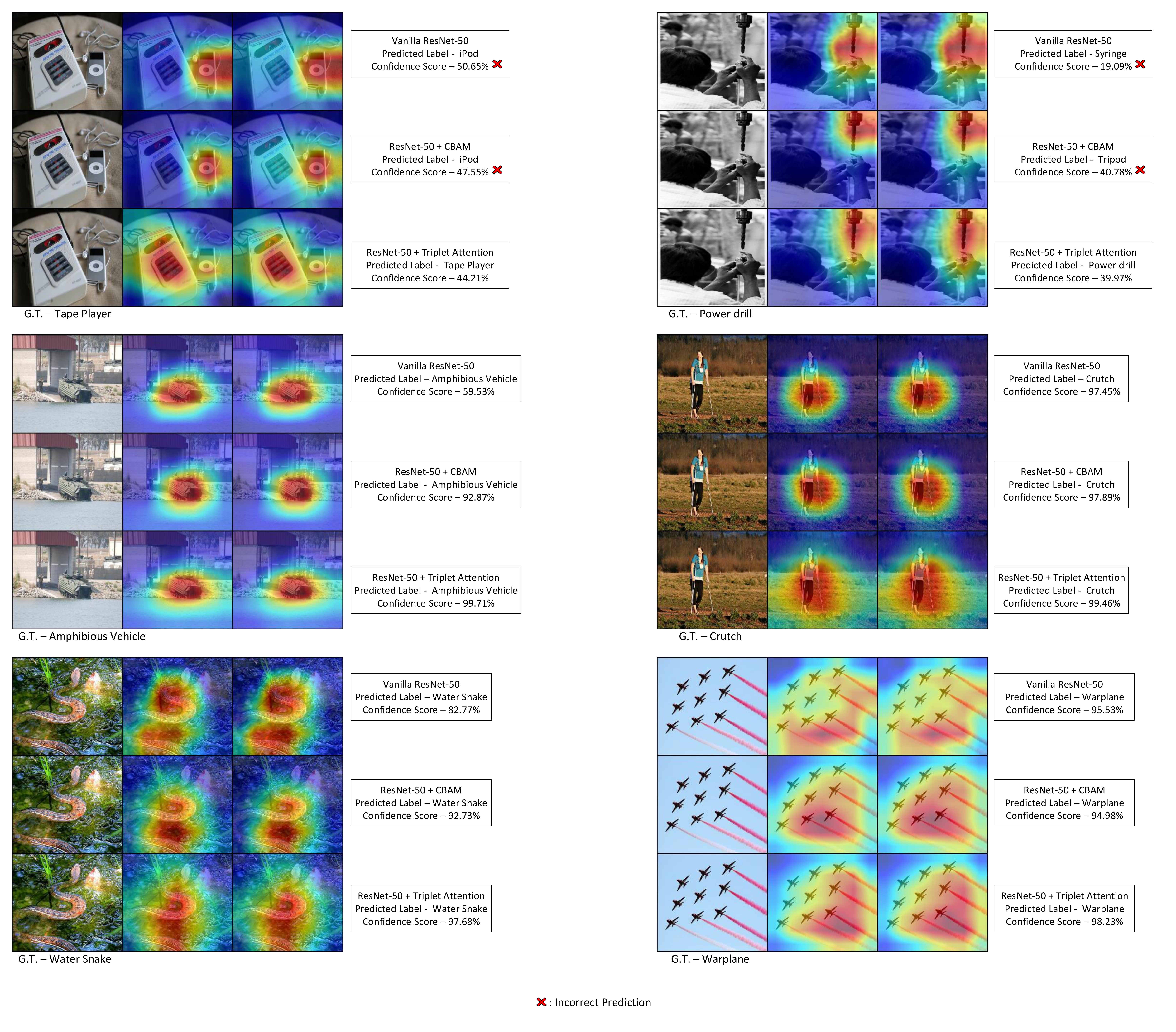} 
  \caption{\textbf{Visualization of GradCAM and GradCAM++ results.} The results were obtained for six random samples from the ImageNet validation set and were compared for a baseline ResNet-50, CBAM integrated ResNet-50 and a triplet attention integrated ResNet-50 architecture. Ground truth (G.T) labels for the images are provided below the original samples and the networks prediction and confidence scores are provided in the corresponding boxes.}
  \label{fig:gradcam1}
\end{figure*}

\begin{table*}[t]
\small
\setlength\tabcolsep{4mm}
\renewcommand{\arraystretch}{1.2}
\begin{center}
\begin{tabular}{|l|c|c|c|c|c|c|c|}
\hline
Backbone & Detectors & AP & AP$_{50}$ & AP$_{75}$ & AP$_{S}$ & AP$_{M}$ & AP$_{L}$\\
\hline\hline
ResNet-50 \cite{he2016deep} & \multirow{4}{*}{Mask RCNN \cite{he2017mask}} & 34.2 & 55.9 & 36.2 & \textbf{18.2} & 37.5 & 46.3\\
ResNet-50 + 1 NL block \cite{wang2018non} & & 34.7 & 56.7 & 36.6 & - & - & - \\
GCNet \cite{gao2019global} & & 35.7 & \textbf{58.4} & 37.6 & - & - & - \\
ResNet-50 + Triplet Attention (Ours) & &  \textbf{35.8} & 57.8 & \textbf{38.1}  & 18.0 & \textbf{38.1} & \textbf{50.7} \\
\hline
\end{tabular}
\end{center}
\caption{\textbf{Instance Segmentation mAP (\%) on MS-COCO }: Triplet Attention results in higher performance gain with minimal computational overhead}
\label{tab:instance}
\end{table*}

\begin{table*}[t]
\small
\setlength\tabcolsep{4mm}
\renewcommand{\arraystretch}{1.2}
\begin{center}
\begin{tabular}{|l|c|c|c|c|c|c|}
\hline
Backbone & Detectors & AP & AP$_{50}$ & AP$_{75}$ & AP$_{M}$ & AP$_{L}$\\
\hline\hline
ResNet-50 \cite{he2016deep} & \multirow{3}{*}{Keypoint RCNN} & 63.9 & 86.4 & 69.3 & 59.4 & 72.4\\
ResNet-50 + CBAM \cite{Woo_2018_ECCV} & & \textbf{64.8} & 85.5 & \textbf{70.9} & \textbf{60.3} & 72.8 \\
ResNet-50 + Triplet Attention (Ours) & & 64.7 & \textbf{85.9} & 70.4 & \textbf{60.3} & \textbf{73.1} \\
\hline
\end{tabular}
\end{center}
\caption{\textbf{Person Keypoints Detection baselines}: Triplet Attention provides improvement over vanilla architecture and competitive results as compared to the more complex CBAM incorporated model.}
\label{tab:keypoint}
\end{table*}

\begin{table*}[t!]
\small
\setlength\tabcolsep{4mm}
\renewcommand{\arraystretch}{1.2}
\begin{center}
\begin{tabular}{|l|c|c|c|c|c|c|c|}
\hline
Backbone & Detectors & AP & AP$_{50}$ & AP$_{75}$ & AP$_{S}$ & AP$_{M}$ & AP$_{L}$\\
\hline\hline
ResNet-50 \cite{he2016deep} & \multirow{3}{*}{Keypoint RCNN} & 53.6 & 82.2 & 58.1 & 36 & 61.4 & 70.8 \\
ResNet-50 + CBAM \cite{Woo_2018_ECCV} & & 54.3 & 82.2 & 59.3 & 37.1 & \textbf{61.9} & 71.4 \\
ResNet-50 + Triplet Attention (Ours) & &  \textbf{54.8} & \textbf{83.1} & \textbf{59.9}  & \textbf{37.4} & \textbf{61.9} & \textbf{72.1} \\
\hline
\end{tabular}
\end{center}
\caption{\textbf{Object detection mAP($\%$) on the MS COCO validation set using the Keypoint RCNN}. Triplet Attention results in consistent higher performance gains across all the metrics.}
\label{tab:bbox}
\end{table*}

\section{GradCAM} \label{grad1}

In addition to the GradCAM results presented in the paper, we observed many more instances of triplet attention generating heatmaps that are consistently tighter or wider when required and more meaningful. We use the same method that we followed in the paper to obtain GradCAM \cite{selvaraju2017grad} and GradCAM++ \cite{chattopadhay2018grad} heatmap visualizations for the ImageNet \cite{deng2009imagenet} test set images that we illustrate in Fig. \ref{fig:gradcam1}.

The most interesting visualization is in the first example (left image on the first row). The image shows two devices - one that resembles a cassette player and an iPod. While this image could potentially benefit from multiple labels and bounding boxes, the class prescribed by the ImageNet dataset is "TapePlayer" (predicted correctly by triplet attention) and not "iPod" (the top class prediction from both CBAM and the vanilla ResNet50). We speculate that the attention maps in triplet attention help the model develop an accurate estimate of global, long-range dependencies in the image. Since the iPod is smaller, its distinct circular control pad coupled with the locality of the discrete convolution operator employed by the ResNet architecture could potentially mislead the network toward predicting the smaller, more recognizable object.

The second example (right image on the first row) also demonstrates an incorrect class prediction that can be attributed to an inability to capture global features. All models focus on a similar region of the image, but CBAM and vanilla ResNet predict the wrong class with reasonably high accuracy. Predicting \textit{power drill} correctly for this image likely requires a representation of the global context since there seem to be few local features that can be associated with that class label.
The other heatmaps continue to suggest that triplet attention produces tighter and more discriminative bounds when appropriate, across a variety of image classes.

\section{COCO Instance Segmentation}

The Mask RCNN architecture introduced in \cite{he2017mask} produces segmentation masks in addition to bounding boxes. We use the Mask RCNN model augmented with our triplet attention layer, trained on the COCO 2017 dataset (as described in section 4.3 of the main paper) to perform instance segmentation, using the detectron2 code base \cite{wu2019detectron2}. We provide our results of various AP scores in Tab. \ref{tab:instance} along with results from other models that used similar training schemes.
On instance segmentation, triplet attention continues to provide a substantial improvement (nearly a 6\% increase across AP scores at negligible computational overhead) over the baseline ResNet50 model and also outperforms other newer, larger models like GCNet \cite{cao2019gcnet}.

\section{COCO Keypoint Detection}

In addition to the other COCO segmentation and object detection tasks, we further train the Mask RCNN model on the COCO human keypoint detection task.
The training configuration is similar to that we used for our Mask RCNN model on the instance segmentation and object detection tasks - we use the same 1x training schedule with identical values for batch size, learning rate, et cetera. as we did for our Mask RCNN model as well as the baseline \cite{he2017mask}. For the keypoint detection head, the model generates 1500 proposals per image using the region proposal network implemented in Faster RCNN \cite{ren2015faster}, which is implemented as the default configuration in detectron2 \cite{wu2019detectron2}.

We provide a table of results comparing our Mask RCNN based keypoint detector to the baseline implementation as well as CBAM \cite{Woo_2018_ECCV}, another method that computationally much more expensive yet obtains similar results. Tab. \ref{tab:keypoint} provides the resulting AP scores for the keypoint annotations on the COCO 2017 validation set. Tab. \ref{tab:bbox} provides the AP scores for the bounding box annotations, which we generate while training on the keypoint annotations.

\end{document}